\RequirePackage{fix-cm}
\documentclass[twocolumn]{svjour3}          
\usepackage[a4paper,margin=1.6cm]{geometry}

\smartqed  
\usepackage{graphicx}
\usepackage{subcaption}
\usepackage{amsmath} 
\usepackage{float}
\usepackage{dblfloatfix}

\usepackage[hyphens]{url}
\usepackage{hyperref}
\usepackage{xurl}
\journalname{Signal Image and Video Processing}
\begin{document}

\title{AI for Mycetoma Diagnosis in Histopathological Images: The MICCAI 2024 Challenge
}


\author{
Hyam Omar Ali \and Sahar Alhesseen \and
Lamis Elkheir \and Adrian Galdran \and
Ming Feng \and Zhixiang Xiong \and
Zengming Lin \and Kele Xu \and
Liang Hu \and Benjamin Keel \and Oliver Mills \and
James Battye \and Akshay Kumar \and
Asra Aslam \and Prasad Dutande \and
Ujjwal Baid \and Bhakti Baheti \and
Suhas Gajre \and Aravind Shrenivas Murali \and
Eung-Joo Lee \and Ahmed Fahal \and Rachid Jennane }

\institute{
H. O. Ali \and R. Jennane \at
Institute Denis Poisson, Orleans University, 45067 Orleans, France \\
\email{hyam.omar-abbass-ali@univ-orleans.fr}
\and
H. O. Ali \and S. Alhesseen \and L. Elkheir \and A. Fahal \at
University of Khartoum, Khartoum, Sudan
\and
A. Galdran \at
TECNALIA, Basque Research and Technology Alliance (BRTA), Bizkaia, Spain 
University of the Basque Country (UPV/EHU), Leioa, Spain
Ikerbasque: Basque Foundation for Science, Bilbao, Spain
\and
M. Feng \and Z. Xiong \and L. Hu \at
Tongji University, Shanghai, China
\and
Z. Lin \and K. Xu \at
National University of Defense Technology, Changsha, China
\and
B. Keel \and O. Mills \and J. Battye \and A. Kumar \at
University of Leeds, Leeds, UK
\and
A. Aslam \at
University of Sheffield, Sheffield, UK
\and
P. Dutande \and S. Gajre \at
SGGS Institute of Engineering and Technology, Nanded, India
\and
U. Baid \at
Georgia Institute of Technology and Emory University, Atlanta, GA, USA
\and 
U. Baid \and B. Baheti \at
Indiana University, Bloomington, USA
\and
A. S. Murali \and E.-J. Lee \at
University of Arizona, Tucson, USA
}

\maketitle

\begin{abstract}
Mycetoma is a neglected tropical disease caused by fungi (Eumycetoma) or bacteria (Actinomycetoma) that can lead to severe tissue destruction, disability, and socioeconomic burden, particularly in resource-limited endemic regions. Accurate diagnosis remains challenging because histopathological interpretation requires experienced pathologists who are often unavailable in these settings. To accelerate the development of artificial intelligence (AI) solutions for mycetoma diagnosis, the Mycetoma MicroImage: Detect and Classify Challenge (mAIcetoma) was organized in conjunction with the 27th International Conference on Medical Image Computing and Computer-Assisted Interventions (MICCAI 2024). The challenge provided a standardized histopathological image dataset (MyData) and comprised two tasks: automated mycetoma grain segmentation and disease classification into eumycetoma and actinomycetoma. Five finalist teams developed diverse deep learning approaches, including ensemble learning, cascaded segmentation-classification frameworks, multi-task learning, attention mechanisms, and self-configuring segmentation networks. All finalist methods achieved strong segmentation performance, while classification results demonstrated the feasibility of accurate automated differentiation between the two disease types. Beyond reporting the challenge outcomes, this paper provides a comparative analysis of the finalist methodologies, summarizes the key lessons learned, and establishes a benchmark to guide future research on AI-assisted diagnosis of mycetoma and other neglected tropical diseases.

\end{abstract}

\section{Introduction}\label{sec1}
Mycetoma is a neglected tropical disease (NTD) that exerts a profound and far-reaching impact on individuals, families, communities, and health systems, particularly in endemic regions \cite{fahal2022mycetoma,zijlstra2016mycetoma}. It is a chronic and progressive disease, characterised by slow-growing lesions that can lead to severe deformities, disability, and social stigma, and compound the suffering of those affected \cite{Fahal_bjs,fahal2004mycetoma}. Despite its classification as an NTD, the burden of mycetoma is disproportionately high in many low-resource settings, especially in countries like Sudan \cite{fahal2022mycetoma,alhaj2024epidemiological}. Although mycetoma was officially declared a health priority by the WHO in 2016 due to its devastating consequences on public health, yet, there is no genuine progress in the management of the affected patients and communities \cite{WHO2025mycetoma,hay2015mycetoma}.

Mycetoma is caused by a wide variety of microorganisms, including more than 70 causative organisms, which can be broadly classified into two categories: Actinomycetoma (AM), caused by bacterial species and Eumycetoma (EM), caused by fungi \cite{Guerrant2011,van2024updated}. The infection primarily occurs when microorganisms from the environment penetrate the skin, often through minor injuries. The chronic granulomatous inflammation caused by these organisms results in localised swelling, sinus tract formation, and the presence of grains in the discharge, which are a diagnostic hallmark of mycetoma \cite{fahal2018mycetoma,williams2013bailey}. 

The disease's insidious onset and slow progression often lead to a delayed diagnosis and subsequent treatment, exacerbating the physical and psychological toll on patients \cite{ahmed2004mycetoma}. This delay, combined with the limited access to adequate healthcare facilities in many endemic regions, creates significant barriers to timely and effective intervention \cite{hay2021mycetoma}. Mycetoma patients frequently experience severe disability due to the destruction of soft tissues, bones, and joints, which can lead to amputation in advanced cases \cite{abbas2018disabling,omer2016hand}. The affected individuals often face social exclusion, economic hardship, and limited access to education and employment due to the debilitating nature of the disease \cite{fahal2024mycetoma}.

The limited availability and accuracy of diagnostic tools in many endemic regions hamper effective diagnosis and management \cite{ahmed2017mycetoma,van2014merits}. In many endemic regions, the histopathological technique remains the only available diagnostic tool, as more advanced diagnostic methods, such as molecular testing or imaging techniques, may be inaccessible due to high costs or limited infrastructure \cite{ahmed1999development,bahar2021mycetoma}. However, its effectiveness is highly dependent on the skill of a well-trained histopathologist, which is often unavailable in rural or resource-limited areas. Furthermore, the complexity of histopathological interpretation poses a challenge for less experienced practitioners, underscoring the need for more robust diagnostic solutions \cite{yousif2010new}. One potential advancement is the development of automated AI-based diagnostic tools, which could assist in the interpretation of histopathological findings. AI-driven systems hold promise for improving diagnostic accuracy and accessibility, particularly in regions where trained specialists are scarce. Such tools could facilitate early diagnosis and ensure that patients receive the appropriate treatment in a timely manner, thereby reducing the burden of mycetoma on affected individuals and health systems alike \cite{ali2024use}. 

In 2023, the first machine learning framework for the semi-automated analysis of mycetoma histopathological images was introduced, achieving a classification accuracy of 91.89\% for distinguishing eumycetoma from actinomycetoma \cite{omar2024evaluation}. Although this study demonstrated the feasibility of applying AI to mycetoma diagnosis, it was limited to a single methodology and did not provide a standardized benchmark for evaluating alternative approaches.

To accelerate research in this emerging field, we organized the Mycetoma MicroImage: Detect and Classify Challenge (mAIcetoma) in conjunction with the 27th International Conference on Medical Image Computing and Computer-Assisted Interventions (MICCAI 2024). The challenge was designed to stimulate the development and objective evaluation of robust AI methods for automated mycetoma grain segmentation and disease classification using histopathological images. Unlike individual methodological studies, the challenge provided a common dataset, standardized evaluation framework, and independent benchmarking environment in which multiple research groups developed and evaluated their approaches under identical conditions.

This paper presents a comprehensive overview of the mAIcetoma Challenge and it is organized as follows: Section~\ref{sec:challenge} describes the design of the mAIcetoma Challenge and the MyData dataset used in the competition in Section~\ref{sec:data}. Section~\ref{sec:methods} presents the methodologies proposed by the finalist teams. Section~\ref{sec:evaluation} introduces the evaluation protocol and ranking scheme adopted in the challenge. Section~\ref{sec:results} presents and analyzes the challenge results, followed by a comparative analysis of the finalist methods. Section~\ref{sec:discussion} discusses the key methodological insights, lessons learned, and limitations. Finally, Section~\ref{sec:conclusion} concludes the paper.

\section{Challenge Design}\label{sec:challenge}
The mAIcetoma Challenge was organized in conjunction with MICCAI 2024 to promote the development of automated methods for detecting and classifying mycetoma grains from histopathological images \cite{Mycetoma2025Challenge}. Beyond providing a competitive platform, the challenge established the first international benchmark for AI-assisted histopathological mycetoma diagnosis, enabling the standardized evaluation and comparison of different methodological approaches using a common dataset and evaluation framework. By bringing together researchers from the medical imaging and AI communities, the challenge aimed to stimulate research on this NTD and establish a reference point for future developments in automated mycetoma diagnosis.

The challenge consisted of two core tasks designed to reflect the routine histopathological diagnostic workflow.

\noindent \textbf{Task 1: Mycetoma Grain Segmentation}\newline
Participants were required to develop algorithms to segment mycetoma grains within histopathological images. Accurate grain segmentation constitutes the first step in routine histopathological assessment because the identification of mycetoma grains are essential before species classification can be performed. The challenge dataset provided expert-annotated grain masks, and participants' results were evaluated based on segmentation performance. Developed algorithms should output the boundary definitions of each detected mycetoma grain within the images. 
Although the challenge title refers to detection, participants were required to produce pixel-wise segmentation masks of mycetoma grains. Therefore, throughout this paper, Task 1 is referred to as the segmentation task.

\begin{figure*}[h!]
	\centering
	\begin{subfigure}[b]{0.23\textwidth}
		\centering
		\includegraphics[width=\textwidth]{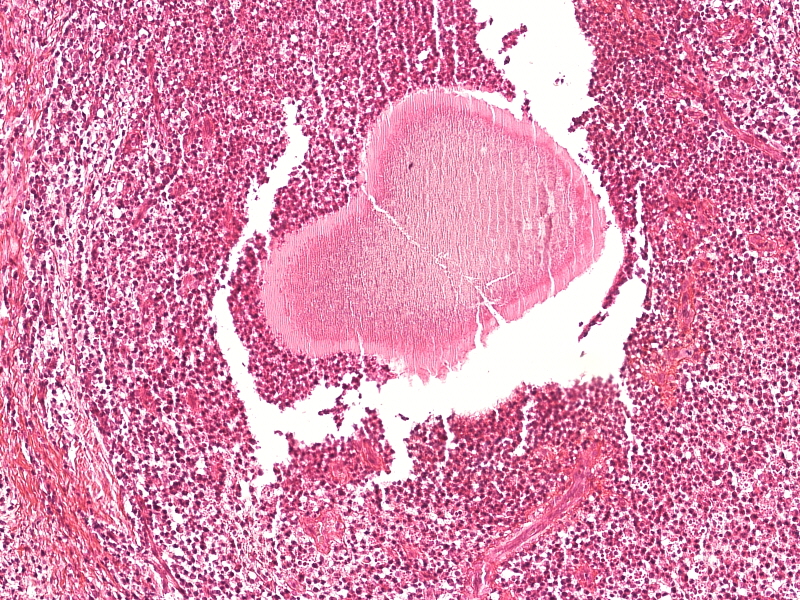} 
		\caption{AM image}
		\label{fig:am_histo}
	\end{subfigure}
	\hfill 
    \begin{subfigure}[b]{0.23\textwidth}
		\centering
		\includegraphics[width=\textwidth]{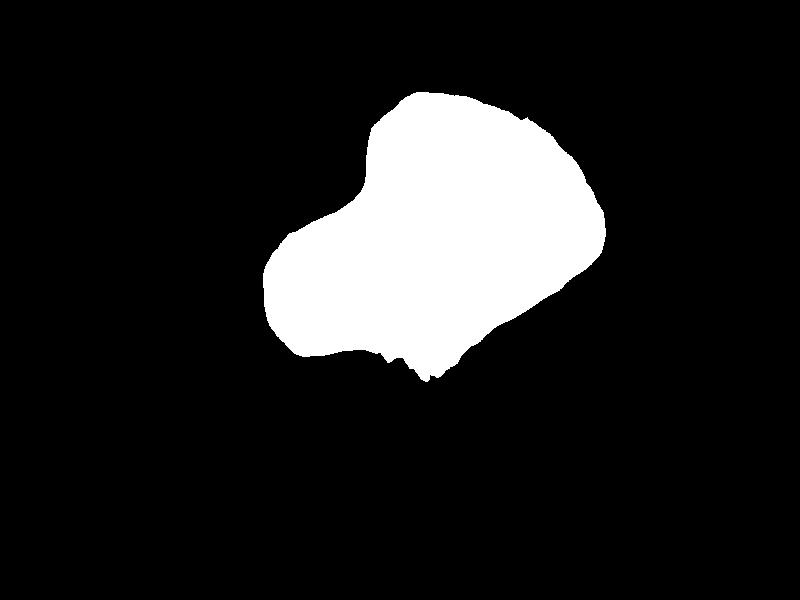} 
		\caption{AM Segmentation Mask}
		\label{fig:am_mask}
	\end{subfigure}
	\hfill
	\begin{subfigure}[b]{0.23\textwidth}
		\centering
		\includegraphics[width=\textwidth]{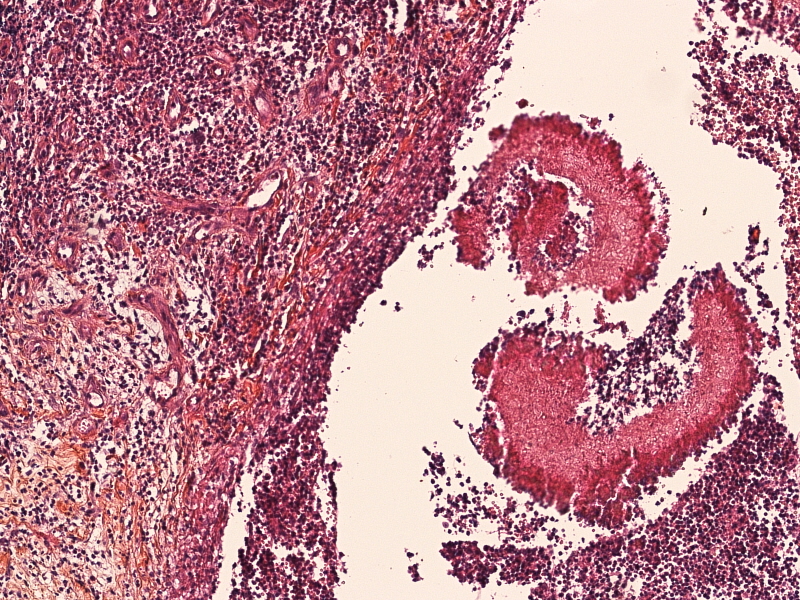}
		\caption{EM image}
		\label{fig:em_histo}
	\end{subfigure}
	\hfill 
	\begin{subfigure}[b]{0.23\textwidth}
		\centering
		\includegraphics[width=\textwidth]{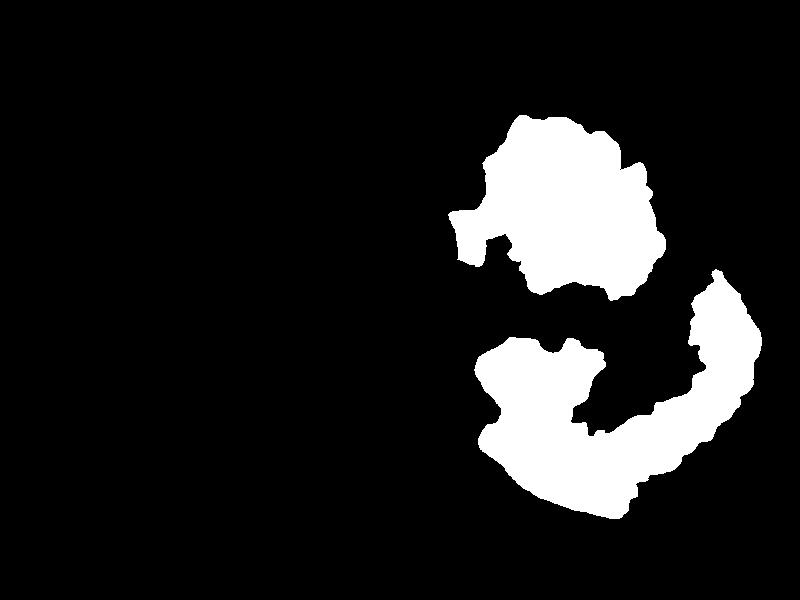}
		\caption{EM Segmentation Mask}
		\label{fig:em_mask}
	\end{subfigure}
		\caption{Representative histopathological images and corresponding segmentation masks}
	\label{fig:mycetoma_images}
\end{figure*}

\noindent{\textbf{Task 2: Mycetoma Type Classification}}\newline
Following grain segmentation, the second task focused on classifying each detected grain as either AM or EM. Accurate classification is critical for determining treatment, as bacterial infections are treated with antibiotics, while fungal infections often require antifungal therapy and/or surgical intervention. For each detected mycetoma grain, the algorithm was supposed to give a classification label indicating whether the detected grain is AM or EM. Fig.\ref{fig:mycetoma_images} illustrates typical histopathological images of both classes and their segmentation masks.

Collectively, these two tasks reproduce the routine diagnostic workflow followed by pathologists, where mycetoma grains are first localized within the tissue section and subsequently classified as AM or EM to support treatment planning.

\subsection{Challenge statistics}
\begin{figure}[h!]
	\centering
	\includegraphics[width=\columnwidth]{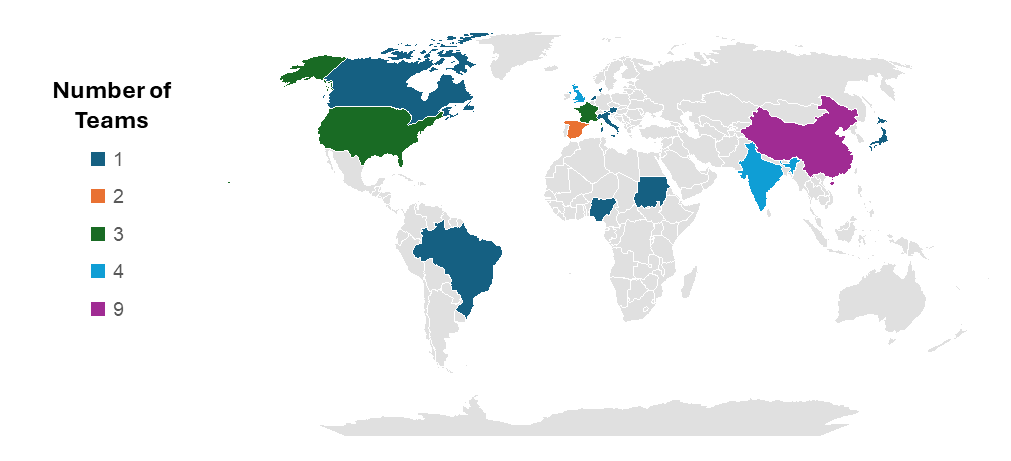}
	\caption{Geographical distribution of participating teams}
	\label{fig:map}
\end{figure}

The challenge attracted multiple international teams from diverse backgrounds, including computer vision researchers, AI developers, and medical imaging experts. A total of 34 teams, each comprising one to five members, registered for the challenge from 16 different countries, Fig.\ref{fig:map}. The participating teams represented a range of experience levels in medical image analysis; 20.59\% were beginners, 64.71\% were intermediate, and 14.71\% were advanced. 

Of the 34 registered teams, five successfully completed the challenge and submitted final solutions for evaluation (Table \ref{Table:demographic}). Three of the teams presented their results in person on the day of the challenge, while the other two teams sent recordings of their presentations.

\begin{table*}[h!]
	\caption{Finalist team demographics}\label{Table:demographic}
	\centering
	\begin{tabular}{|l|c|l|l|}	
		\hline
		Team Name      & Number of Team Members & Country        & Level of Experience \\
		\hline
		Adrian & 1                      & Spain          & Advanced \\
		Macaron        & 5                      & China          & Intermediate \\
		Minions        & 5                      & United Kingdom & Intermediate \\
		Tiger      & 4                      & United States  & Intermediate \\
		VSI            & 2                      & United States  & Intermediate \\
	
		\hline
	\end{tabular}
	\end{table*}

   \subsection{Challenge Workflow and Timeline}
The challenge followed a well-defined structure, including multiple phases to ensure fair evaluation and participant engagement (Table \ref{Table:phases}). 
Participants were required to submit their results through an online submission portal, ensuring a blind evaluation where test set ground truth labels were not available to teams. This protocol ensured that the reported performance reflected the generalization ability of the submitted methods rather than optimization to the test set.
\begin{table*}[h!]
	\caption{Timeline of the mAIcetoma Challenge}\label{Table:phases}
	\centering
	\begin{tabular}{|p{3cm}|p{6cm}|p{2cm}|}
		\hline
		\textbf{Phase} & \textbf{Description} & \textbf{Date} \\
		\hline
		Challenge announcement and registration opening & 
		Participants register and gain access to labelled training and validation data. & 
		May 2, 2024 \\
		\hline
		Submission portal opening & 
		Participants submit predictions on the test dataset. & 
		August 19, 2024 \\
		\hline
		Submission Deadline & 
		All final results must be submitted for evaluation. & 
		September 6, 2024 \\
		\hline
		Evaluation \& Results Announcement & 
		The challenge team evaluates the submissions and announces the results to participants. & 
		September 23, 2024 \\
		\hline
		MICCAI 2024 Presentation & 
		The teams' rankings were revealed, and the teams presented their approaches at the challenge session. & 
		October 6, 2024 \\
		\hline
	\end{tabular}
\end{table*}

\section{Dataset Description}\label{sec:data}
A novel histopathological image dataset for mycetoma, MyData, was developed specifically to support automated mycetoma grain segmentation and classification \cite{ali2024mydata}. The dataset comprises surgical biopsies collected over the past five years from patients diagnosed with different clinical presentations and causative agents of mycetoma at the Mycetoma Research Centre (MRC), Sudan. Ethical approval for data collection and use was obtained from the Soba University Hospital Ethical Committee, Khartoum, Sudan.

Following surgical excision, tissue blocks were prepared and sectioned into two to three histological slides per specimen. The sections were stained with haematoxylin and eosin (H\&E) and digitized using a Nikon Eclipse 80i optical microscope. Histopathological images were acquired in RGB colour space at $10 \times$ magnification, which corresponds to the routine magnification used by pathologists and provides a field of view that typically captures complete mycetoma grains together with the surrounding tissue architecture required for both segmentation and classification. Higher magnifications such as  $20 \times$ or  $40 \times$ primarily provide additional cellular detail while substantially reducing the field of view; however, cellular-level morphology is not required for mycetoma diagnosis, which relies principally on the grains and their relationship with the surrounding tissue.

An average of six images were acquired per patient in the original MyData database, resulting in a total of 864 histopathological images from 142 patients \cite{ali2024mydata}. Each image was manually annotated by expert pathologists with a binary segmentation mask delineating the mycetoma grains, providing the ground truth for the segmentation task. For the classification task, each image was assigned a diagnostic label according to the corresponding expert pathological diagnosis.

For the mAIcetoma Challenge, a subset of this database was released as the official challenge dataset. The released challenge dataset comprised 766 histopathological images from 127 patients, including 415 EM images from 70 patients and 351 AM images from 57 patients. This dataset was divided into training, validation, and test sets using a patient-wise splitting strategy specifically designed to prevent data leakage and ensure a realistic assessment of model generalization. Consequently, all images from a given patient were assigned exclusively to a single subset. The final split consisted of 89 training patients (593 images, 70.1\%), 23 validation patients (91 images, 18.1\%), and 15 testing patients (82 images, 11.8\%). This strategy ensured that no patient contributed images to more than one subset while maintaining a representative distribution of both mycetoma classes across the training, validation, and test sets (Table~\ref{tab:dataset_split}).
\begin{table*}[h!]
	\caption{Distribution of the challenge dataset }
	\label{tab:dataset_split}
	\centering
	\begin{tabular}{l|c|c|c|c|c|c}
		\hline
		\textbf{Dataset split} & \textbf{BM patients} & \textbf{FM patients} & \textbf{Total patients (\%)} & \textbf{AM images} & \textbf{EM images} & \textbf{Total images} \\
		\hline
		Training   & 43 & 46 & 89 (70.1\%)  & 282 & 311 & 593 \\
		Validation &  9 & 14 & 23 (18.1\%)  &  38 &  53 &  91 \\
		Test       &  5 & 10 & 15 (11.8\%)  &  31 &  51 &  82 \\
		\hline
		\textbf{Total} & \textbf{57} & \textbf{70} & \textbf{127 (100\%)} & \textbf{351} & \textbf{415} & \textbf{766} \\
		\hline
	\end{tabular}
\end{table*}

\section{Participating Teams and Methods}\label{sec:methods}
\subsection{Methods}
This section summarizes the methodologies proposed by the five finalist teams for the segmentation and classification tasks.
\begin{enumerate}
	\item Adrian\newline
	The Adrian team adopted a data-centric approach, beginning with dataset preprocessing to improve the quality of the training data. Semi-duplicate images were identified by computing the Structural Similarity Index Measure (SSIM) between all image pairs. The image pairs were ranked from the highest to the lowest similarity scores and the most similar pairs were visually inspected. Images representing the same visual content were subsequently merged, improving the consistency of the ground-truth annotations. An overview of the complete Adrian framework is shown in Figure~\ref{fig:adrian_architecture}.
	\begin{figure}[h!]
		\centering
		\includegraphics[width=\columnwidth]{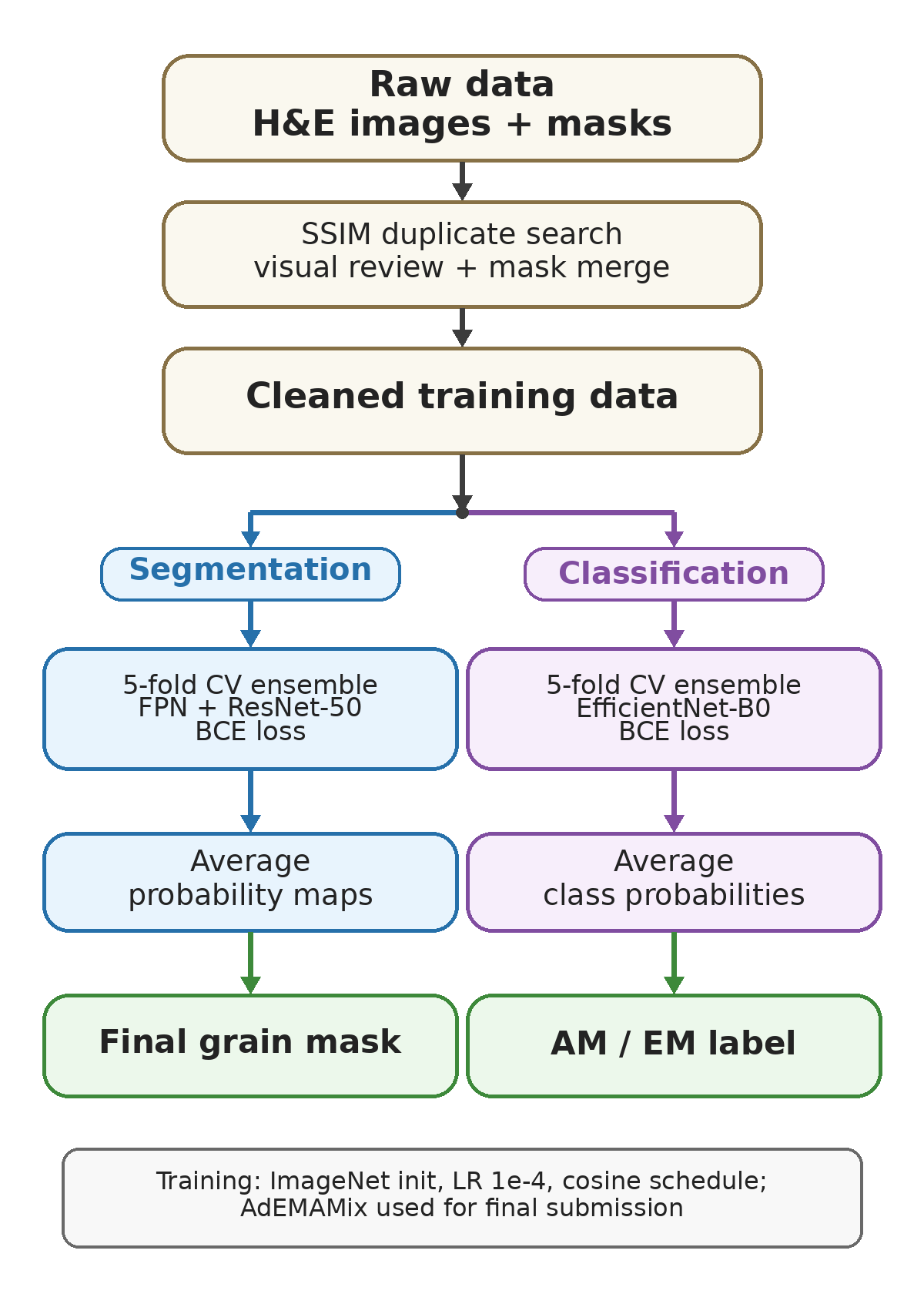}
		\caption{Overview of Adrian's framework}
		\label{fig:adrian_architecture}
	\end{figure}
		
	For the segmentation task, the team employed a five-fold cross-validation ensemble based on a Feature Pyramid Network (FPN) \cite{lin2017feature} with a ResNet-50 encoder pretrained on ImageNet (encoder depth of 5). The decoder consisted of 256 pyramid channels and 128 segmentation channels with a dropout rate of 0.2. A Mix Vision Transformer (MiT) \cite{chen2022transmix} decoder was adopted with embedding dimensions of [64, 128, 320, 512], depths of [3, 8, 27, 3], and approximately 60 million trainable parameters. Five models were trained by rotating the validation fold, and the final segmentation was obtained by averaging their predictions. Binary cross-entropy was used as the optimization loss.
	
	For the classification task, the team adopted the same five-fold cross-validation ensemble strategy using EfficientNet-B0 \cite{tan2019efficientnet}. Both segmentation and classification models were optimized using the AdEMAMix optimizer with a learning rate of $10^{-4}$ and a cosine annealing learning-rate scheduler over 20 training cycles of five epochs each. The team initially compared AdamW with the recently proposed AdEMAMix optimizer \cite{pagliardini2024ademamix}. Although both optimizers achieved competitive results, AdEMAMix consistently produced slightly better performance with reduced overfitting and was therefore adopted for the final submission.
	
	A distinguishing feature of the Adrian approach was its emphasis on data-centric model development. According to the authors, careful dataset cleaning prior to training, combined with ensemble learning and robust optimization, contributed substantially to the strong performance achieved in both challenge tasks.

	\item Macaroon\newline
The Macaroon team proposed a cascaded framework comprising separate segmentation and classification models for mycetoma grain segmentation and classification (Fig.~\ref{fig:macroon}). The segmentation model first generated a probability map, which was converted into a binary mask using Otsu's automatic thresholding algorithm. Ensemble predictions were obtained by averaging the probability maps of four segmentation models (U-Net, U-Net++, DeepLabV3, and DeepLabV3+) before thresholding. The resulting binary mask was then used to generate the region of interest (ROI) for the subsequent classification stage.

ROI generation was based directly on the binary segmentation mask obtained after Otsu thresholding, without any additional thresholding or filtering. The largest connected component was extracted, and its minimum enclosing rectangle was computed to define the lesion location. The bounding box was subsequently expanded by 20 pixels in all directions to preserve surrounding contextual information before being resized to $512 \times 512$ pixels for classification. During inference, segmentation was performed using a sliding-window strategy with a window size of $512 \times 512$ pixels and a stride of 256 pixels, corresponding to a 50\% overlap between adjacent windows. Otsu thresholding was the only post-processing step applied to the segmentation output.

For classification, the team employed an ensemble of four ImageNet-pretrained models consisting of ResNet, Xception, EfficientNet-B0, and InceptionV3. Similar to the segmentation task, the final prediction was obtained by averaging the outputs of the four classifiers. Both segmentation and classification models were trained using the official challenge data split, the Adam optimizer, and a batch size of 16. Early stopping was employed with a patience of five epochs to reduce overfitting.

Data augmentation was applied independently to the two tasks. For segmentation, augmentations included Resize, HorizontalFlip, VerticalFlip, RandomRotate90, and RandomResizedCrop. For classification, RandomCrop, HorizontalFlip, VerticalFlip, GaussianBlur, RandomRotate90, RandomBrightnessContrast, and ShiftScaleRotate were used. The augmentation parameters were specified as follows: RandomCrop with a crop ratio of 0.8-1.0; RandomRotate90 using rotations of $0^\circ$, $90^\circ$, $180^\circ$, and $270^\circ$ with probability 0.5; RandomBrightnessContrast with brightness and contrast adjustments within $\pm0.2$ (probability 0.5); ShiftScaleRotate with shift and scale ranges of $\pm10\%$, rotation range of $\pm15^\circ$ (probability 0.5); RandomResizedCrop with a scale range of 0.7-1.0 and aspect ratio of 0.9-1.1; and GaussianBlur with probability 0.3, kernel sizes between 3 and 7, and $\sigma$ values ranging from 0.1 to 2.0.

The key methodological contribution of the Macaroon approach was the cascaded segmentation--classification framework combined with ensemble learning. Ensemble learning has consistently been shown to improve model robustness and generalization \cite{ganaie2022ensemble}. The authors argued that directly classifying the full histopathological image introduces substantial background noise because the lesion occupies only a small portion of the image. By first localizing the lesion and subsequently classifying an ROI centred on the segmented grain, the framework focused the classifier on diagnostically relevant tissue while preserving local contextual information. According to the authors, the combination of ROI-based learning, ensemble models, and extensive data augmentation contributed to the strong performance of their method, whereas the principal limitation was the potential propagation of segmentation errors to the classification stage.
\begin{figure}[t]
	\centering
	\includegraphics[width=\columnwidth]{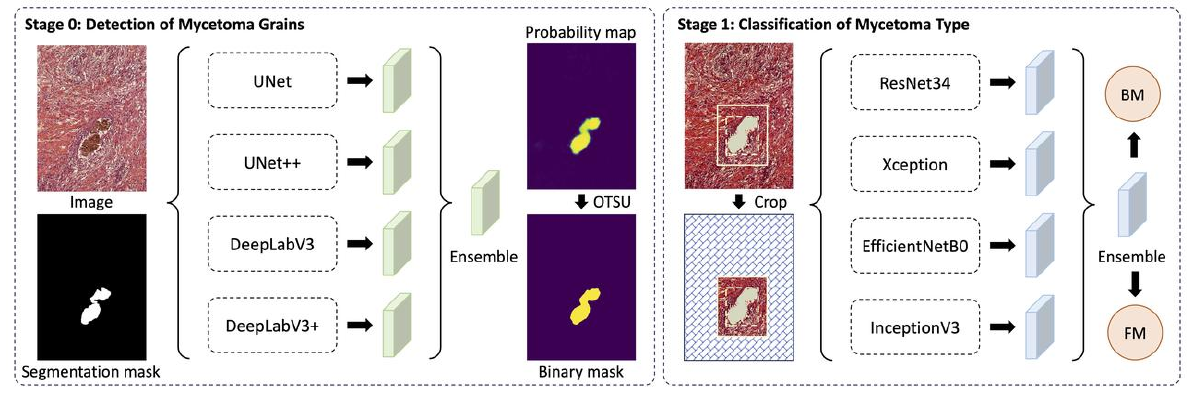}
	\caption{Overview of the Macaroon framework}
	\label{fig:macroon}
\end{figure}

	\item Minions\newline
	The Minions approach consisted of multiple stages including data preprocessing, segmentation, classification, hyperparameter optimization, and post-processing, as illustrated in Fig.~\ref{fig:min1}. The team submitted two classification models. The first incorporated clinical variables, handcrafted features, and the predicted segmentation masks, whereas the second served as a baseline model using only the RGB histopathological images.
	\begin{figure}[h!]
		\centering
		\includegraphics[width=\columnwidth]{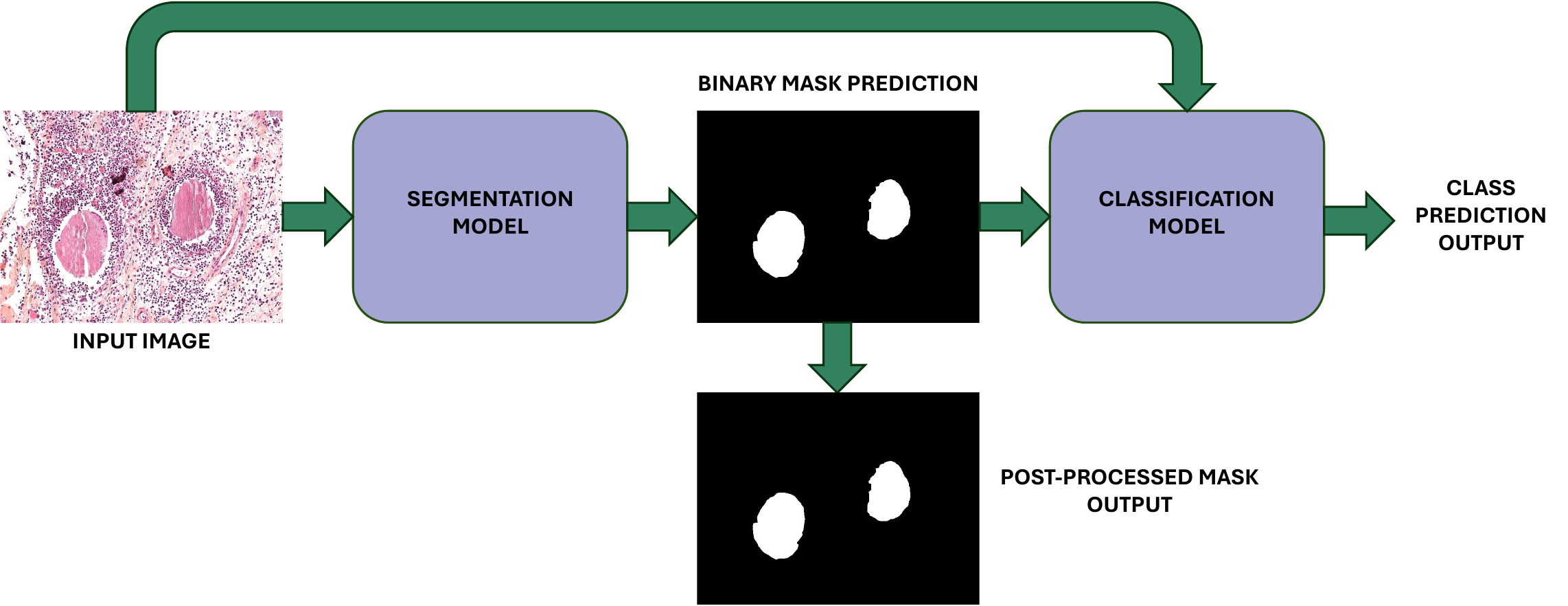}
		\caption{Overview of the Minions data processing pipeline}
		\label{fig:min1}
	\end{figure}
	
	Data preprocessing involved extensive quality control of the training and validation datasets. A small number of masks stored as three-channel images were converted to single-channel binary masks to ensure consistency. Duplicate images containing different grain annotations were merged into a single ground-truth mask. In addition, overlapping but non-identical images were identified using ORB feature extraction and the OpenCV BFMatcher algorithm. A translation vector was estimated to align image pairs, and masks were merged when the overlap exceeded 90\%. Finally, image--mask pairs containing large unlabelled grains were removed. This cleaning process improved the validation Dice coefficient from 0.78 to 0.83 before hyperparameter optimization, highlighting the importance of data quality.
	
	A modified U-Net architecture was employed for mycetoma grain segmentation (Fig.~\ref{fig:min2}). The network consisted of three encoder stages with 8, 16, and 32 convolutional filters, followed by a 64-filter bottleneck and a symmetric decoder with skip connections. In addition to the segmentation head, a classification head attached to the bottleneck enabled multi-task learning by jointly optimizing segmentation and classification objectives. The total loss was defined as
	$\mathrm{Loss}=\mathrm{SegmentationLoss}+\alpha\times\mathrm{ClassificationLoss}$,
	where the segmentation loss combined binary cross-entropy and Dice loss, while the classification loss used binary cross-entropy. Hyperparameter optimization selected a batch size of 8, learning rate of $2\times10^{-4}$, and $\alpha=0.2$. Following inference, binary masks were generated using a threshold of 0.5 and refined through morphological opening and closing operations, connected-component analysis, and hole filling to improve segmentation quality.
	\begin{figure}[h!]
		\centering
		\includegraphics[width=\columnwidth]{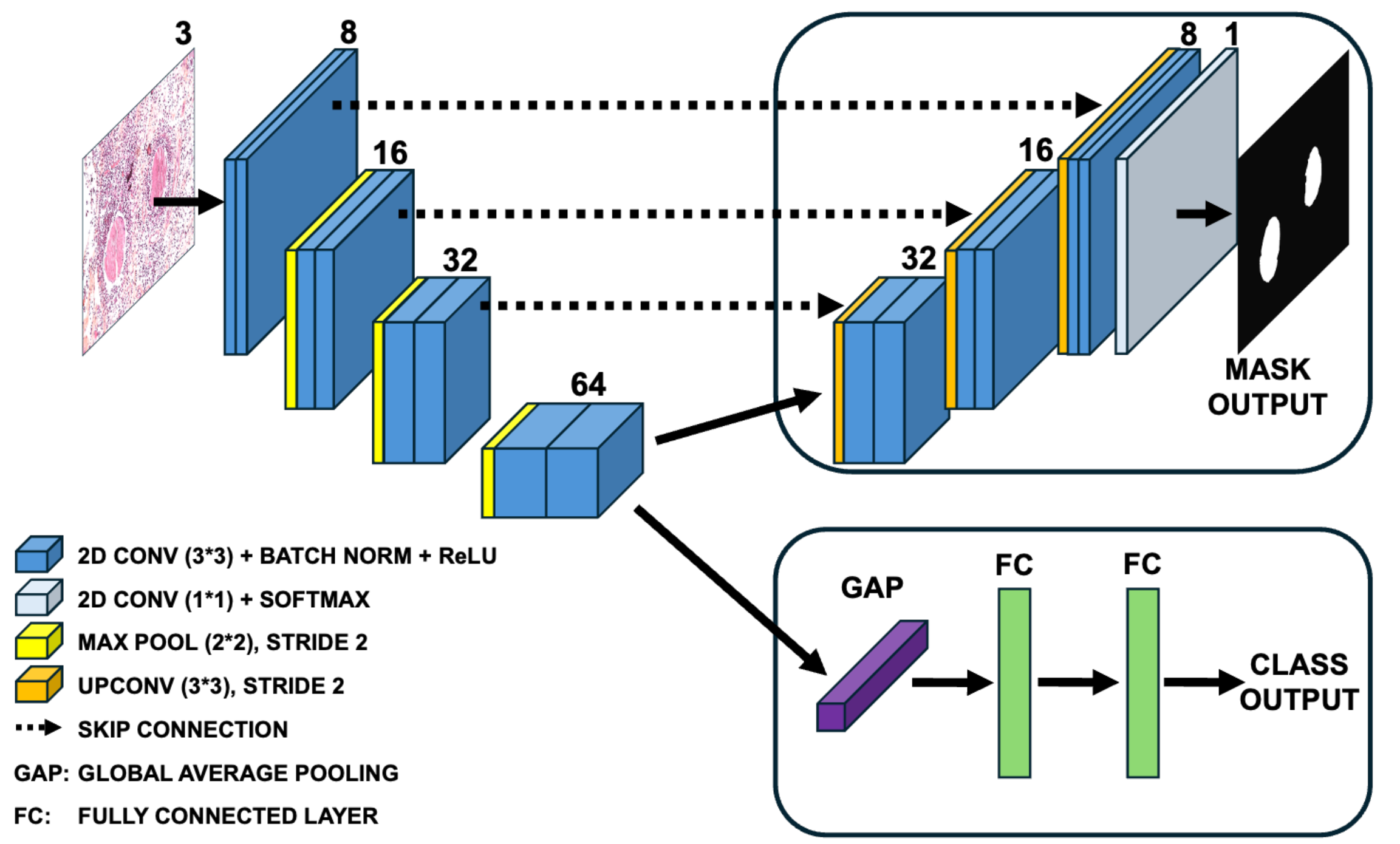}
		\caption{Minions segmentation architecture}
		\label{fig:min2}
	\end{figure}
	
	The classification model employed a DenseNet-169 backbone with ImageNet-pretrained weights implemented using MONAI \cite{cardoso2022monai}, followed by a two-layer multilayer perceptron classifier (Fig.~\ref{fig:min3}). Two classification experiments were performed. The first incorporated the predicted segmentation masks together with 27 handcrafted features describing grain morphology, whereas the second used only the RGB images as input. The handcrafted features included nine colour features (mean, variance, and skewness for each RGB channel), seven texture features extracted from Gray Level Co-occurrence Matrices (GLCM) and Local Binary Patterns (LBP), and eleven shape descriptors characterizing grain morphology and boundary irregularity. The GLCM features were computed using a distance of one pixel and a horizontal ($0^\circ$) orientation, while the extracted statistics included contrast, homogeneity, and energy. LBP features were computed using the uniform operator with a radius of one pixel and eight neighbouring points, from which the mean, variance, skewness, and kurtosis of the LBP histogram were extracted. All handcrafted features were normalized to the range [0,1] before being concatenated with the DenseNet feature representation.
	\begin{figure}[h!]
		\centering
		\includegraphics[width=\columnwidth]{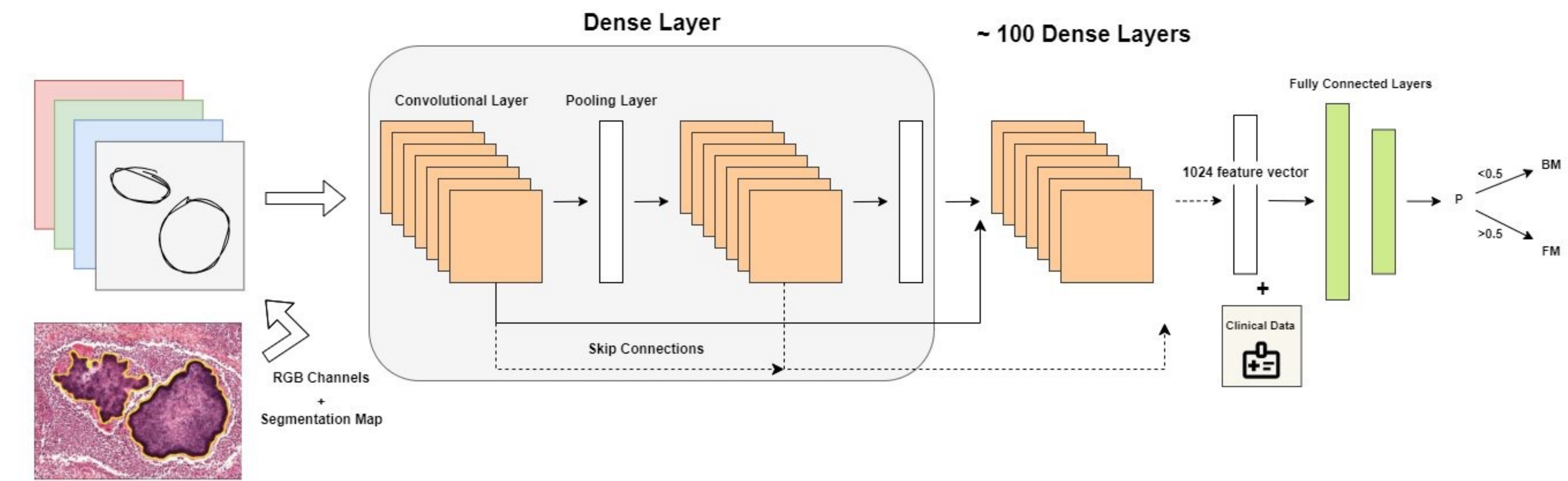}
		\caption{Minions classification architecture}
		\label{fig:min3}
	\end{figure}
	
	The classifier was trained using binary cross-entropy loss and the Adam optimizer. Data augmentation included random horizontal and vertical flips (probability 0.8) together with colour jitter (brightness, contrast, and saturation = 0.2; hue = 0.02). Bayesian optimization was used to tune the learning rate ($10^{-5}$--$10^{-2}$), batch size (8--32), and weight decay ($10^{-4}$--$10^{-2}$). The optimal hyperparameters were a learning rate of $4\times10^{-4}$, batch size of 12, and weight decay of $10^{-3}$. The baseline model achieved superior validation performance and was therefore selected for the final submission, with only eight of the 167 test predictions falling within the uncertain probability range (0.25--0.75), suggesting that such cases could be referred for expert pathological review.
	\item Tiger\newline
	The Tiger team proposed a two-stage framework consisting of separate segmentation and classification models. The segmentation model (Fig.~\ref{fig:tiger_seg}) was based on DeepLabV3+ with a shared ResNet50 encoder and was designed to localize mycetoma grain regions at the pixel level. Histopathological images were resized to $224 \times 224$ pixels before being processed by the network. Low-level encoder features preserved boundary and spatial information, while high-level semantic features were processed by an Atrous Spatial Pyramid Pooling (ASPP) module to capture multi-scale contextual information, allowing the model to accommodate the variable size and morphology of mycetoma grains. The ASPP features were fused with low-level skip connections and decoded using three parallel attention-guided decoder branches incorporating CBAM, spatial attention, and channel attention modules, respectively. The final segmentation mask was obtained using either the selected decoder output or the average fusion of the three decoder predictions. The model was trained using the Adam optimizer with Dice loss and evaluated using the Dice score.
	\begin{figure}[h!]
		\centering
		\includegraphics[height=0.4\textheight,keepaspectratio]{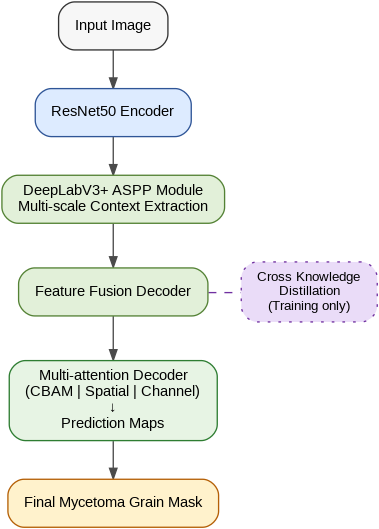}
		\caption{Overview of Tiger's segmentation framework}
		\label{fig:tiger_seg}
	\end{figure}
	
	\begin{figure}[h!]
		\centering
		\includegraphics[height=0.55\textheight,keepaspectratio]{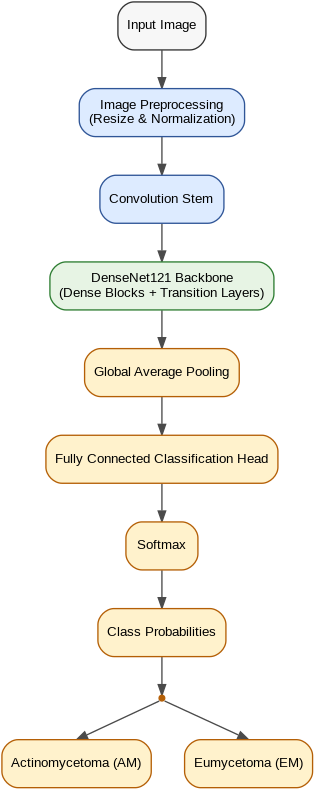}
		\caption{Overview of Tiger's classification framework}
		\label{fig:tiger_cls}
	\end{figure}
	
	A key component of the segmentation framework was the Cross Knowledge Distillation (CKD) module, which operated only during training. Rather than transferring knowledge between segmentation and classification, CKD acted as a cross-branch regularization mechanism in which the three decoder branches exchanged soft segmentation probability maps. This encouraged the branches to learn complementary representations while maintaining prediction consistency, particularly around ambiguous lesion boundaries. During inference, the CKD module was not used and the final prediction was generated solely from the trained segmentation network.
	
	The classification model (Fig.~\ref{fig:tiger_cls}) employed a DenseNet121 classifier initialized with ImageNet-pretrained weights. Images were resized to $224 \times 224$ pixels and normalized before being passed through the network, which consisted of an initial convolutional stem, dense blocks, transition layers, global average pooling, and a fully connected classification head with softmax activation for AM and EM prediction. The model was optimized using Adam with a learning rate of $10^{-4}$ and Cross-Entropy loss for 100 epochs using a batch size of 32. Model selection was based on the best validation Accuracy and AUC. Data augmentation included horizontal and vertical flips, rotations of $90^\circ$, $180^\circ$, and $270^\circ$, together with brightness and contrast variations, while validation images were processed only by resizing and normalization.
	
	The principal strength of the Tiger framework was the combination of multi-scale context extraction, attention-guided decoding, and Cross Knowledge Distillation. The three complementary attention mechanisms enhanced lesion localization by emphasizing diagnostically relevant channels and spatial regions while suppressing background tissue. According to the authors, CKD further improved segmentation by encouraging agreement among the decoder branches through soft prediction alignment, thereby reducing overfitting around uncertain grain boundaries. The classification model complemented the segmentation stage by exploiting the dense feature reuse of DenseNet121 to distinguish bacterial from fungal mycetoma. The authors identified staining variability, irregular grain morphology, fragmented lesions, and the visual similarity between grains and surrounding tissue as the principal challenges encountered during model development.

	\item VSI\newline
	For the segmentation task, the VSI team employed the 2D configuration of nnU-Net \cite{isensee2021nnu}, which automatically configured most of the network architecture for the dataset. The resulting architecture was a PlainConvUNet encoder--decoder with skip connections over eight resolution stages. Each stage consisted of two $3\times3$ convolutional layers followed by instance normalization and LeakyReLU activation (slope = 0.01). The number of feature channels increased from 32 to 512 across the network, and deep supervision was retained throughout training. Since the dataset consisted of RGB histopathological images without meaningful voxel spacing, the default nnU-Net spacing-based resampling was disabled and all images were resized to a fixed resolution of $640 \times 896$ pixels. Training followed the default nnU-Net optimization strategy using stochastic gradient descent with Nesterov momentum (0.99), a learning rate of 0.01 with polynomial decay, and the default compound Dice and cross-entropy loss.
	
	To further improve segmentation quality, the team incorporated a fully connected Conditional Random Field (CRF) \cite{sutton2012introduction} as a post-processing step. The CRF operated on the softmax probability maps using Gaussian spatial and bilateral position--RGB pairwise terms, followed by five mean-field inference iterations. The CRF parameters were selected through a small validation sweep to maximize the Dice coefficient. This post-processing step consistently refined grain boundaries and reduced small false-positive regions.
	
	For the classification task, the team fine-tuned an ImageNet-pretrained ResNet50 \cite{he2016deep} by replacing the original classification head with a single sigmoid output for binary classification. Images were resized to $224 \times 224$ pixels and normalized using the standard ImageNet mean and standard deviation. The model was optimized using Adam with a learning rate of $1\times10^{-4}$ and binary cross-entropy loss. To address the mild class imbalance, class weights were assigned according to inverse class frequency (0.92 for EM and 1.10 for AM). Standard data augmentation, including horizontal and vertical flipping, rotations of approximately $\pm10^\circ$, and brightness/contrast jitter of approximately $\pm10$--20\%, was applied using the default nnU-Net augmentation pipeline.
	
	A key strength of the VSI approach was the use of the self-configuring nnU-Net framework with minimal manual tuning, complemented by CRF-based boundary refinement. According to the authors, correct image normalization, task-specific input resolutions, and model selection based on MCC rather than accuracy were important factors contributing to the performance of their method, while the computational overhead introduced by the CRF and the use of a single segmentation model instead of an ensemble were identified as its principal limitations.

\end{enumerate}

\section{Evaluation method}\label{sec:evaluation}
The evaluation framework was designed to provide a comprehensive and clinically meaningful assessment of participants' algorithms. Rather than relying on a single performance metric, multiple complementary metrics were used to evaluate different aspects of algorithm performance. 

For the classification task, performance was evaluated using sensitivity, specificity, accuracy, and Matthews Correlation Coefficient (MCC) \cite{boughorbel2017optimal}. Sensitivity and specificity quantify the ability to correctly identify each disease class while minimizing false-negative and false-positive classifications, respectively, whereas accuracy provides an overall measure of classification performance. MCC was included because it considers all four elements of the confusion matrix and provides a robust measure of performance for moderately imbalanced datasets.

For the segmentation task, sensitivity, specificity, accuracy, and Dice coefficient \cite{setiawan2020image} were used. Sensitivity measures the correct segmentation of mycetoma grain pixels, specificity evaluates the ability to avoid incorrectly labelling surrounding tissue as grain, accuracy reflects overall pixel-wise correctness, and the Dice coefficient quantifies the spatial overlap between the predicted and reference masks, making it one of the most widely adopted metrics for biomedical image segmentation.

As part of the challenge protocol, default scores were predefined for incomplete submissions to ensure a consistent evaluation procedure. However, all finalist teams submitted complete results; therefore, this rule did not influence the final rankings.

\subsection{Ranking Scheme}
The final ranking was determined using a weighted-sum approach for both the classification and segmentation tasks. The evaluation protocol, including the selected metrics and their corresponding weights, was defined before the challenge by the organizing committee in consultation with clinicians from the MRC, ensuring that the ranking reflected clinically relevant priorities while providing a balanced and transparent comparison of all submissions.

For the classification task, the moderately imbalanced class distribution motivated assigning the highest weight to MCC (40\%), as it provides a more reliable summary of classification performance than accuracy alone. Accuracy received the second-highest weight (30\%), while sensitivity and specificity were assigned equal weights of 15\% each to ensure balanced evaluation of both disease classes.

For the segmentation task, equal weights of 25\% were assigned to sensitivity, specificity, accuracy, and Dice coefficient because each metric captures a complementary aspect of segmentation quality, including grain localisation, avoidance of false-positive segmentation, overall pixel-wise accuracy, and agreement between the predicted and reference masks.

The overall ranking scores for the classification and segmentation tasks were computed using Eqs.~(\ref{eq:classification}) and (\ref{eq:segmentation}), respectively. Sen, Spec, Acc, MCC, and Dice denote sensitivity, specificity, accuracy, MCC, and Dice coefficient, respectively, while $w_{\mathrm{sen}}$, $w_{\mathrm{spec}}$, $w_{\mathrm{acc}}$, $w_{\mathrm{MCC}}$, and $w_{\mathrm{Dice}}$ denote their corresponding weights.
\begin{align}
	\mathrm{Score}_{\mathrm{cls}}
	&=
	w_{\mathrm{sen}}\mathrm{Sen}
	+w_{\mathrm{spec}}\mathrm{Spec}
	+w_{\mathrm{acc}}\mathrm{Acc}
	+w_{\mathrm{MCC}}\mathrm{MCC},
	\label{eq:classification}\\
	\mathrm{Score}_{\mathrm{seg}}
	&=
	w_{\mathrm{sen}}\mathrm{Sen}
	+w_{\mathrm{spec}}\mathrm{Spec}
	+w_{\mathrm{acc}}\mathrm{Acc}
	+w_{\mathrm{Dice}}\mathrm{Dice}.
	\label{eq:segmentation}
\end{align}

Finally, submissions were ranked according to their weighted scores in descending order, with the highest score assigned the top rank. This approach provides a simple, transparent, and interpretable framework that balances multiple clinically relevant performance measures while reflecting the overall quality of each submission.

\section{Results}\label{sec:results}
\subsection{Comparative Analysis of the Finalist Methods}\label{sec:Comparative}
Although all finalist teams addressed the same segmentation and classification tasks, they adopted markedly different methodological strategies. Table~\ref{Table:comparison} summarizes the principal design philosophies of the five approaches, highlighting their principal methodological innovations, strengths, and potential limitations. Rather than comparing individual implementation details, this comparison emphasizes the broader methodological trends that emerged across the challenge and provides context for interpreting the quantitative results presented in the following subsections.

\begin{table*}[h!]
	\caption{Comparative analysis of the finalist approaches.}
	\label{Table:comparison}
	\centering
	\begin{tabular}{|p{1.8cm}|p{2.8cm}|p{3.8cm}|p{3.6cm}|p{3.8cm}|}
		\hline
		\textbf{Team} & \textbf{Core Strategy} & \textbf{Distinctive Feature} & \textbf{Primary Strength} & \textbf{Key Insight} \\
		\hline
		
		Adrian &
		Data-centric ensemble learning &
		Dataset cleaning using SSIM followed by transformer-based segmentation and EfficientNet-B0 ensemble classification &
		Improved robustness through careful data cleaning, ensemble learning, and strong optimization &
		Careful data preparation combined with ensemble learning can substantially improve robustness \\
		\hline
				
		Macaroon &
		Cascaded segmentation-classification &
		ROI extraction from segmentation results followed by ensemble classification &
		Reduces background interference by focusing classification on diagnostically relevant regions &
		ROI-guided cascaded pipelines effectively reduce background interference during classification \\
		\hline
		
		Minions &
		Hybrid multi-task learning &
		Joint segmentation-classification framework combined with handcrafted morphological features and extensive data cleaning &
		Careful data inspection, annotation refinement, and robust deep learning achieved competitive performance &
		Correcting annotation inconsistencies improved performance, while handcrafted features were largely redundant \\
		\hline

		Tiger &
		Attention-guided deep learning &
		Cross Knowledge Distillation with multi-attention decoder branches &
		Rich multi-scale feature learning and improved boundary localisation through attention mechanisms &
		Attention mechanisms and knowledge distillation enhance feature representation for complex histopathological images \\
		\hline
		
		VSI &
		Self-configuring biomedical segmentation &
		2D nnU-Net with CRF-based post-processing and a ResNet50 classifier &
		Strong baseline performance with minimal manual architecture design and stable training &
		Self-configuring frameworks provide strong biomedical imaging baselines with minimal manual tuning \\
		\hline

	\end{tabular}
\end{table*}

Despite the methodological diversity of the finalist approaches, several common themes emerged. First, all teams relied on deep learning architectures but combined them with complementary strategies such as data-centric preprocessing, ensemble learning, attention mechanisms, multi-task optimization, or cascaded segmentation-classification frameworks. Second, several of the highest-performing methods emphasized careful data preparation, localization of diagnostically relevant regions before classification, or ensemble learning, suggesting that overall pipeline design is as important as the underlying network architecture. Finally, the diversity of competitive solutions demonstrates that no single architectural paradigm dominates this task, providing valuable insights for future research on AI-assisted histopathological diagnosis of mycetoma and other NTDs.

\subsection{Challenge Results}

The performance of the finalist teams was evaluated separately for the segmentation and classification tasks using the evaluation metrics and ranking scheme described in Section~\ref{sec:evaluation}. The quantitative results for both tasks are presented below.

The segmentation task evaluated each method's ability to accurately segment mycetoma grains within histopathological images. Table~\ref{Table:seg} summarizes the segmentation results achieved by the finalist teams. Adrian obtained the highest weighted score (93.56\%), followed closely by Macaroon (92.52\%) and Tiger (91.98\%). Notably, all finalist teams achieved weighted scores exceeding 91\%, demonstrating consistently strong segmentation performance despite employing different methodological strategies.

\begin{table*}[h!]
	\caption{Segmentation results}\label{Table:seg}
	\centering
	
	\begin{tabular}{|l|c|c|c|c|c|}
		\hline
		\textbf{Team} & \textbf{Sensitivity} & \textbf{Specificity} & \textbf{Accuracy} & \textbf{Dice} & \textbf{Weighted Sum} \\
		\hline
		Adrian   & $0.9914$ & $0.8885$ & $0.9806$  & $0.8820$  & $93.56\%$  \\
		\hline
		Macaroon & $0.9740$ & $0.9124$ & $0.9706$  & $0.8439$  & $92.52\%$  \\
		\hline
		Tiger    & $0.9956$ & $0.8406$ & $0.9786$  & $0.8646$  & $91.98\%$  \\
		\hline
		Minions  & $0.9863$ & $0.9863$ & $0.9785$  & $0.8362$  & $91.42\%$  \\
		\hline
		VSI      & $0.9834$ & $0.8696$ & $0.9723$  & $0.8295$  & $91.37\%$  \\
		\hline
	\end{tabular}
	
\end{table*}

For the classification task, the models were evaluated according to their ability to distinguish between EM and AM. Table~\ref{Table:class} summarizes the classification results. Adrian again achieved the highest weighted score (96.14\%), followed by Macaroon (93.11\%) and Minions (92.06\%). Compared with segmentation, greater variability was observed among the classification results, indicating that differentiating between the two mycetoma types remains the more challenging task. The inclusion of MCC in the ranking scheme ensured a balanced evaluation by accounting for the moderate class imbalance.

\begin{table*}[h!]
	\caption{Classification results}\label{Table:class}
	\centering
	\begin{tabular}{|l|c|c|c|c|c|}
		\hline
		\textbf{Team} & \textbf{Sensitivity} & \textbf{Specificity} & \textbf{Accuracy} & \textbf{MCC} & \textbf{Weighted Sum} \\
		\hline
		Adrian   & $0.9659$ & $0.9828$ & $0.9726$  & $0.9435$  & $96.14\%$  \\
		\hline
		Macaroon & $0.9545$ & $0.9483$ & $0.9521$  & $0.9003$  & $93.11\%$  \\
		\hline
		Minions  & $0.9545$ & $0.9310$ & $0.9452$  & $0.8856$  & $92.06\%$  \\
		\hline
		VSI      & $0.9091$ & $0.9310$ & $0.9178$  & $0.8317$  & $88.40\%$  \\
		\hline
		Tiger    & $0.8068$ & $0.9655$ & $0.8699$  & $0.7559$  & $82.91\%$  \\
		\hline

	\end{tabular}
\end{table*}

Overall, Adrian achieved the strongest performance across both segmentation and classification tasks, followed by Macaroon, while Minions and Tiger also demonstrated highly competitive performance. Interestingly, these leading teams adopted markedly different methodological strategies. Adrian combined data-centric preprocessing with ensemble learning, Macaroon employed a cascaded segmentation-classification framework based on ROI extraction, Minions integrated multi-task learning with handcrafted morphological features, whereas Tiger relied on attention mechanisms and Cross Knowledge Distillation. The challenge results demonstrate that multiple methodological strategies can achieve highly competitive performance for automated mycetoma grain segmentation and disease classification. The consistently strong performance achieved across the finalist teams establishes a robust benchmark for future research on AI-assisted mycetoma diagnosis.

\section{Discussion}\label{sec:discussion}
\subsection{Key Lessons Learned}
One of the most important findings of the mAIcetoma challenge was the consistently high segmentation performance achieved by all finalist teams despite their markedly different methodological approaches. All five teams achieved weighted segmentation scores exceeding 91\%, demonstrating that automated localisation of mycetoma grains can be performed reliably using a variety of modern deep learning architectures. This observation supports the challenge design of separating grain segmentation from disease classification, as accurate localisation of diagnostically relevant regions remains clinically valuable even when disease subtype classification is uncertain and requires expert review. In contrast, the greater variability observed in the classification task suggests that distinguishing between EM and AM remains the more challenging component of the diagnostic pipeline.

A recurring theme among the finalist teams was the critical importance of data quality and preprocessing. Both the Adrian and Minions teams independently emphasized that careful visual inspection of the images and ground-truth masks enabled the identification and correction of annotation inconsistencies and duplicated data, leading to improved model performance. These observations demonstrate that careful dataset curation can be as important as model architecture in achieving robust performance. Similarly, the VSI team highlighted the importance of standardizing image size and intensity normalization to ensure stable model training. Collectively, these observations suggest that data-centric strategies can contribute as much to model performance as architectural innovations. The Minions team further observed that images exhibiting intermediate RGB intensity distributions were more frequently misclassified, indicating that staining variability remains an important source of error. Consequently, preprocessing techniques such as colour normalization, histogram scaling, and data augmentation should be considered in future model development.

The challenge also demonstrated that multiple methodological paradigms can achieve competitive performance. Rather than relying on a single dominant architecture, the finalist teams successfully adopted diverse strategies, including ensemble learning, multi-task learning, cascaded segmentation-classification frameworks, attention mechanisms, Cross Knowledge Distillation, and self-configuring segmentation networks. The Macaroon team emphasized the effectiveness of ROI-based cascaded learning in reducing background interference during classification, whereas the Tiger team highlighted the benefits of attention mechanisms and Cross Knowledge Distillation for improving feature representation and segmentation precision. Together with the comparative analysis presented in Section~\ref{sec:Comparative}, these findings indicate that overall pipeline design, rather than the choice of a single network architecture, plays a central role in achieving robust diagnostic performance.

Another important lesson concerned model complexity and generalization. The Adrian team reported that smaller classification networks can achieve performance comparable to larger models while reducing the risk of overfitting. Likewise, the VSI team observed that self-configuring frameworks such as nnU-Net simplify model development but require careful consideration of computational efficiency, particularly when computationally intensive post-processing techniques such as Conditional Random Fields are employed. The Minions team further demonstrated that incorporating handcrafted morphological descriptors did not improve performance over their baseline deep learning model, indicating that the learned image representations were sufficient for this task. Together with their emphasis on identifying and correcting annotation inconsistencies, these findings further reinforce the importance of data quality over additional feature engineering.

Finally, several teams emphasized the importance of incorporating domain knowledge and evaluating model robustness beyond the challenge dataset. The Macaroon team noted that a deeper understanding of the underlying pathology could facilitate more effective model design, while the Minions team recommended external validation using independent cohorts to better assess generalizability. Collectively, these observations highlight several important directions for future research, including improved integration of clinical expertise, development of computationally efficient models, and validation across diverse patient populations and imaging conditions.

\subsection{General Discussion}
The mAIcetoma challenge demonstrated the feasibility of applying AI to the automated detection and classification of mycetoma from histopathological images. The consistently strong performance achieved by the finalist teams, despite their diverse methodological approaches, highlights the potential of deep learning to support mycetoma diagnosis and assist pathologists in clinical decision-making, particularly in low-resource settings where access to expert pathology services may be limited. Rather than identifying a single optimal architecture, the challenge demonstrated that successful AI-assisted diagnosis depends on the integration of high-quality data, appropriate preprocessing, robust model design, and carefully designed evaluation protocols.

At the same time, the challenge highlighted several limitations that should be addressed in future iterations. The current dataset does not include complementary clinical or demographic information that could further improve diagnostic performance and enable more comprehensive clinical decision support. Variability in staining also remains an important source of heterogeneity and may affect model generalizability if not adequately addressed through preprocessing or data augmentation. In addition, the annotations were generated by a single expert pathologist. Although this ensured annotation consistency, future datasets would benefit from multiple independent expert annotations to assess inter- and intra-observer variability and further strengthen the reference standard. Expanding the dataset to include samples collected from multiple centres, together with additional diseases that share similar histopathological characteristics with mycetoma, would further improve model robustness and clinical applicability.

The evaluation framework adopted standard performance metrics that are widely used in medical image analysis. Sensitivity, specificity, accuracy, and MCC were used to evaluate classification performance, while segmentation performance was assessed using sensitivity, specificity, accuracy, and the Dice coefficient. These metrics provided a comprehensive and balanced assessment of model performance and enabled fair comparison across submissions. Future editions of the challenge may benefit from incorporating additional evaluation criteria, such as the F1-score for classification and analyses of annotation variability, to provide complementary perspectives on model performance. Furthermore, the observations reported by several finalist teams reinforce the importance of preprocessing techniques, including colour normalization, data augmentation, and careful quality control of annotations, as integral components of robust AI pipelines rather than optional preprocessing steps.

As the first international challenge dedicated to AI-assisted mycetoma diagnosis, mAIcetoma establishes a valuable benchmark for future research in this NTD. Future work should focus on validating the developed models using independent multi-centre cohorts, investigating emerging AI paradigms such as self-supervised learning, and explainable AI, and developing computationally efficient systems suitable for deployment in routine clinical practice. Ultimately, translating these models into accessible desktop or mobile decision-support tools, together with continued collaboration between AI researchers, clinicians, pathologists, and policymakers, will be essential for ensuring that advances in artificial intelligence lead to meaningful improvements in mycetoma diagnosis and patient care.

\section{Conclusion}\label{sec:conclusion}
This paper presented an overview of the mAIcetoma Challenge held in conjunction with the MICCAI 2024, summarizing the methodologies and outcomes of the finalist teams. The challenge demonstrated the feasibility of applying AI to the automated detection and classification of mycetoma from histopathological images. Despite adopting diverse methodological strategies, all finalist teams achieved highly competitive performance, with Dice coefficients exceeding 0.82 for mycetoma grain segmentation and the best classification model achieving an accuracy of 97.3\%. These findings demonstrate the potential of AI-assisted approaches to support mycetoma diagnosis, particularly in resource-limited settings. Beyond benchmarking model performance, the challenge provided valuable methodological insights into the importance of high-quality data, careful preprocessing, robust model design, and comprehensive evaluation strategies. The diversity of successful approaches further demonstrated that no single architectural paradigm dominates this task, providing a strong foundation for future research on AI-assisted diagnosis of mycetoma and other neglected tropical diseases. Future work will focus on expanding the dataset through multi-centre collaborations, validating the developed models on independent cohorts, and facilitating their integration into clinical workflows. To promote reproducible research and encourage continued development of AI models for mycetoma diagnosis, the challenge dataset remains publicly available to the research community.

%

\section*{Declarations}

\subsection*{Funding}
This research did not receive any specific grant from funding agencies in the public, commercial, or not-for-profit sectors.

\subsection*{Competing Interests}
The authors declare that they have no known competing financial interests or personal relationships that could have appeared to influence the work reported in this paper.

\subsection*{Data Availability}
MyData dataset used in the mAIcetoma Challenge is publicly available through the Zenodo repository (\url{https://zenodo.org/records/13655082}) and the AfricAI repository (\url{https://xnat.bmia.nl/data/projects/africai_miccai2024_maIcetoma}). Additional details regarding the dataset, licensing, and access procedures are provided in \cite{ali2024mydata}.

\subsection*{Code Availability}
Not applicable.

\subsection*{Ethics Approval}
Not applicable.

\subsection*{Consent to Participate}
Not applicable.


\bibliographystyle{spphys}       
\bibliography{sn-bibliography}   

\end{document}